%% file: main.tex
\pdfoutput=1

\documentclass[11pt]{article}

\usepackage[]{EMNLP2022}

\usepackage{times}
\usepackage{latexsym}

\usepackage[T1]{fontenc}

\usepackage[utf8]{inputenc}

\usepackage{microtype}

\usepackage{inconsolata}

\usepackage{ulem}
\usepackage{times}
\usepackage{latexsym}
\usepackage{amsmath}
\usepackage{graphicx}
\usepackage{mathrsfs} 
\usepackage{multirow}
\usepackage{graphicx}
\usepackage{hyperref}
\usepackage[hyphenbreaks]{breakurl}
\usepackage{booktabs}
\usepackage{amssymb}

\PassOptionsToPackage{hyphens}{url}\usepackage{hyperref}

\title{DisCup: Discriminator Cooperative Unlikelihood Prompt-tuning for Controllable Text Generation}

\author{Hanqing Zhang  \\ Beijing Institute of Technology, \\ Beijing, China \\
        \texttt{zhanghanqing@bit.edu.cn}
        \And
        Dawei Song\thanks{ \ \ Corresponding author, also with The Open University, United Kingdom.} \\
        Beijing Institute of Technology, \\ Beijing, China \\
        \texttt{dwsong@bit.edu.cn}}

\begin{document}
\maketitle
\begin{abstract}

Prompt learning with immensely large Casual Language Models (CLMs) has been shown promising for attribute-controllable text generation (CTG). However, vanilla prompt tuning tends to imitate training corpus characteristics beyond the control attributes, resulting in a poor generalization ability. Moreover, it is less able to capture the relationship between different attributes, further limiting the control performance. In this paper, we propose a new CTG approach, namely DisCup, which incorporates the attribute knowledge of discriminator to optimize the control-prompts, steering a frozen CLM to produce attribute-specific texts. Specifically, the frozen CLM model, capable of producing multitudinous texts, is first used to generate the next-token candidates based on the context, so as to ensure the diversity of tokens to be predicted. Then, we leverage an attribute-discriminator to select 
desired/undesired tokens from those candidates, providing the inter-attribute knowledge. Finally, we bridge the above two traits by an unlikelihood objective for prompt-tuning. Extensive experimental results show that DisCup can achieve a new state-of-the-art control performance while maintaining an efficient and high-quality text generation, only relying on around 10 virtual tokens\footnote{The code implementation  is available at: \href{https://github.com/littlehacker26/Discriminator-Cooperative-Unlikelihood-Prompt-Tuning}{https://github.com/littlehacker26/disc-cooperative-up-tuning}}.

\end{abstract}

\input{introduction}

\input{relatedwork}

\input{methodology}

\input{experiment}

\input{conclusion}

\input{limitation}

\normalem
\bibliography{anthology,custom}
\bibliographystyle{acl_natbib}

\clearpage
\thispagestyle{empty}

\input{appendices}

\end{document}

%% file: introduction.tex
\section{Introduction}
Attribute-controllable text generation (CTG) aims to produce texts that satisfy desired attributes (e.g., sentiment, topic, etc.),  facilitating safer and more practical text generation applications. For example, we need to control the emotion and politeness of generated text in a dialogue system for a more friendly interaction, while it is also crucial to avoid generating mindless and offensive content such as racial discrimination and toxic words. Current Transformer-based pre-trained language models (PLMs), especially casual language models (CLMs) like the GPT family~\cite{brown2020language}, have enabled generation of texts of an unprecedented quality. However, due to the lack of interpretability of deep neural networks, it is often hard to guarantee the controllability of these models ~\cite{ctg_survey}. This is a challenging and largely unsolved problem.

The most natural ways of using transformer-based PLMs for CTG are to fine-tune or retrain the models~\cite{chan2021cocon, ctrl,gdc,DART}, so as to control the PLMs to generate texts satisfying the control attributes. Such methods have achieved certain performance breakthroughs in this area.  Nevertheless, the scale of PLMs is getting larger in recent years, making the PLMs resource-intensive to fine-tune or retrain. Therefore, increasing attention has been paid to the decoding-time methods, where a PLM is always fixed and a guided module is used to steer the text generation process.

The core idea of existing decoding-time approaches is  to train an external guided module that consists of a discriminator~\cite{inverse_prompt, yang-klein-2021-fudge,pplm, kumar2021controlled,liu-etal-2020-data} or generative discriminator~\cite{liu-etal-2021-dexperts, krause-etal-2021-gedi-generative}, to adjust the probability of naturally producing a token by the PLM at the decoding phase. This type of method exhibits a strong controllability while maintaining the generalization ability of the original PLM. However, those methods decouple the guided module from the generative PLM, resulting in either computational challenges (i.e., longer inference time or additional parameters) or a negative impact on text generation quality (i.e., arbitrary output text with a lower perplexity).

More recently, CTG approaches based on prompt-tuning are proposed~\cite{li-liang-2021-prefix, tailor_2022}. Control-prompts are usually trained separately with some different attribute-specific corpus, using \textit{Maximum Likelihood Estimation (MLE)} on the traditional next-token prediction task (see more details in Appendix~\ref{mle_detail}). The trained control-prompts are then used as a prefix to steer the attribute-specific generation. This mechanism manifests a promising text quality and a lower computational cost; however, it still suffers from the following drawbacks: (\textit{\textbf{Problem 1}}) The learned control-prompts may absorb the features of training corpus aimlessly and are easily overfitted to other aspects in the training data beyond the control attributes, such as domain style, resulting in the generation of monotonous text. For example, if the sentiment control prompts are optimized in the movie review data, then the generated texts are generally relevant to movies even given prompt text from other domains. The detailed examples are given in Appendix~\ref{vanilla_example}. (\textit{\textbf{Problem 2}}) Attribute control prompts are often trained independently with single-attribute corpus, resulting in the inability to capture the relationship between multiple attributes. However, the previous works~\cite{liu-etal-2021-dexperts,gedi,contrast_prefix} reveal that attribute reference is an important factor in CTG. For example, when generating sentences toward a positive sentiment, we hope the model can refer to the negative ones, so as to achieve a better control performance.

To tackle the aforementioned problems, we propose \textbf{DisCup}, a \textbf{Dis}criminator \textbf{C}ooperative \textbf{U}nlikelihood \textbf{P}rompt-tuning approach for CTG. The key idea is to move the attribute-discriminator, which is usually used to re-rank candidate tokens in the decoding-time approaches, to the training phase for augmenting control-prompt learning, allowing the learned prompts to inherit the advantages of decode-time approaches to some extent. Specifically, the knowledge of the attribute-discriminator is distilled into some continuous control-prompts by a novel objective,  using unlikelihood training~\cite{Welleck2020Neural} to steer a frozen CLM to produce attribute-specific texts. Different from maximizing the probability of the next token under a given context, the proposed  objective contains two parts: a \textit{likelihood objective} that encourages the model to generate candidate tokens satisfying the desired attributes, and an \textit{unlikelihood objective} that allows the model to alleviate generating candidate tokens inconsistent with the target attributes. 

In our approach, the candidate tokens are naturally self-generated by the frozen CLM (i.e., the top-$k$ highly probable tokens under the given context), which is trained on a large corpus and capable of producing multitudinous texts, rather than the ground-truth tokens appearing in the training samples. This will help the control-prompts relieve the problem of over-fitting the features in training data other than target attributes, while ensuring that the CTG model does not deviate far from the original CLM. As such, the aforementioned \textit{\textbf{Problem 1}} can be addressed. Meanwhile, re-ranking the candidate tokens by the attribute-discriminator to chose the desired/undesired tokens, would also indirectly incorporate the relationship between different attributes (thus addressing \textit{\textbf{Problem 2}}), and the unlikelihood training would further enhance the control performance of the CTG model. 

Experimental results on two attribute-controllable generation tasks, i.e., sentiment control and toxicity avoidance, prove that DisCup can achieve a new SOTA control performance. In addition, our method shows a strong comprehension ability, which can simultaneously guarantee a better text quality and a lower computational cost. 

Our main contributions are as follows: (1) We provide a new alternative for controllable generation, by moving the attribute-discriminator in the scheme of decoding-time approaches into the training phase for prompt learning. This allows the model to take advantages of both prompt-tuning and decoding-time methods. (2) We propose a novel unlikelihood training strategy to enhance the control-prompts learning,  which uses the knowledge in the attribute-discriminator to select the likely/unlikely target tokens from the self-generated tokens of the frozen CLM, instead of carrying out the traditional next-token prediction based on a training corpus. (3) We conduct extensive experiments on the tasks of sentiment control and toxicity avoidance, and the results prove the  effectiveness of DisCup and at same time highlight the promise of prompt-learning in CTG.


%% file: relatedwork.tex
\section{Related Work}

A large number of PLM-based CTG approaches have recently emerged.  The most natural way is to ~\textbf{retrain/refactor} the PLMs to establish CTG-specific models. CTRL~\cite{ctrl} is a representative early work, which trains a conditioned language model on the training corpus containing a variety of control code.  \citet{att_alignment} propose an alignment function module to transform the attribute representation, and steer the GPT2-based text generation. CoCon (Content-Conditioner)~\cite{chan2021cocon} injects a control block into the GPT2 model and provides the control code as a separate input, then retrains the whole module by self-supervised learning. \citet{zhang-etal-2020-pointer} propose POINTER, which modifies the structure of Transformer to generate text in a progressive manner for lexically constrained text generation. \citet{metion_flag} propose a Mention Flags (MF) module, which is injected into the decoder of the Transformer, to achieve a higher level of constraint satisfaction.  However, retraining/refactoring the PLMs faces the challenge of lacking labeled data and high computational costs.

As the size of PLMs rapidly increases, the retraining approaches become computationally expensive. Therefore, \textbf{decode-time approaches}, in which the parameters of PLM are fixed, become widely applied to CTG. PPLM~\cite{pplm} uses a simple attribute classifier on the head of a PLM to update the hidden layer by gradient feedback, to achieve the goal of generating desired attribute texts. Instead of updating the hidden layer,  Fudge~\cite{yang-klein-2021-fudge} directly uses the discriminator to select candidate tokens produced by a frozen GPT2 model. In order to accelerate the generation process, GeDi~\cite{gedi} trains the class-conditional language model (CCLM) as generative discriminators to guide the generation from a base GPT2. Plug-and-Blend~\cite{lin2021plug} extends GeDi  to controllable story generation by introducing a planner module. Similarly,  DEXPERT~\cite{liu-etal-2021-dexperts} fine-tunes GPT2 as expert (anti-expert) to re-rank the predictions of the PLM. Decoding-time approaches usually can achieve a competitive control effect, yet fall short in text quality and inference speed, since the guided modules are always decoupled from the generator.

Recently, \textbf{prompt-based methods} for CTG have been proposed. \citet{li-liang-2021-prefix} propose the use of prefix tuning, which freezes the PLM's parameters and back-propagates the error to optimize a continuous task-specific vector to realize a controllable text generation. Based on prefix-tuning, \citet{contrast_prefix} takes into consideration the relationship among prefixes and trains multiple prefixes simultaneously. \citet{tailor_2022} leverage the prompt-tuning~\cite{p_tuning} to establish a multi-attribution control framework, namely Tailor.  Our work shows some similarity to Tailor, but differs in that we focus on improving the inherent problems in the vanilla prompt tuning, instead of prompt-based multi-attribute control. 

To the best of our knowledge, we are the first to  move the discriminator to the training phase, allowing the PLMs to learn a re-ranked token distribution by incorporating the attribute discriminator information, and optimize control-prompts for attribute-specific text generation. As a result, the proposed method inherits the advantages of both decode-time and prompt-tuning approaches to generate higher-quality texts, with a faster inference speed and a better control performance. DIRECTOR~\cite{DIRECTOR} also combines the discriminator in training phase for text generation. However it adds extra classifier architecture to the standard decoder-layer, which deviates from the paradigm of prompt-tuning.

%% file: methodology.tex
\section{Problem Definition}
In this paper, we are concerned with the task of attribute-controllable text generation that aims to steer a casual language model (CLM) (i.e., GPT2, see more details in Appendix~\ref{language_model}) to generate texts satisfying the attribute constraint (e.g., sentiment, topic, toxicity, etc.). Concretely,  given the prompt text $X_{1:t} = \{x_1,x_2, \dots x_t\}$, the goal of our task is to generate the continuations $X_{t+1:n} = \{x_{t+1}, x_{t+2}, \dots, x_n \} $, and let the whole text $X_{1:n}$ conform to the target control attribute denoted as $C$.  It can be formally described as:
\begin{equation}
 P(X|C) = p_{\theta}\left(X_{t+1:n} | X_{1:t}, C \right), 
\end{equation}
 where $C$ represents the control attribute, which may vary according to different tasks.  We expect to establish a CTG model $p_{\theta}(x)$, so that the generated texts $X$ can satisfy the control attribute and have good fluidity and diversity at the same time.


\begin{figure*}[h] 
    \centering 
    \includegraphics[width=0.8\textwidth]{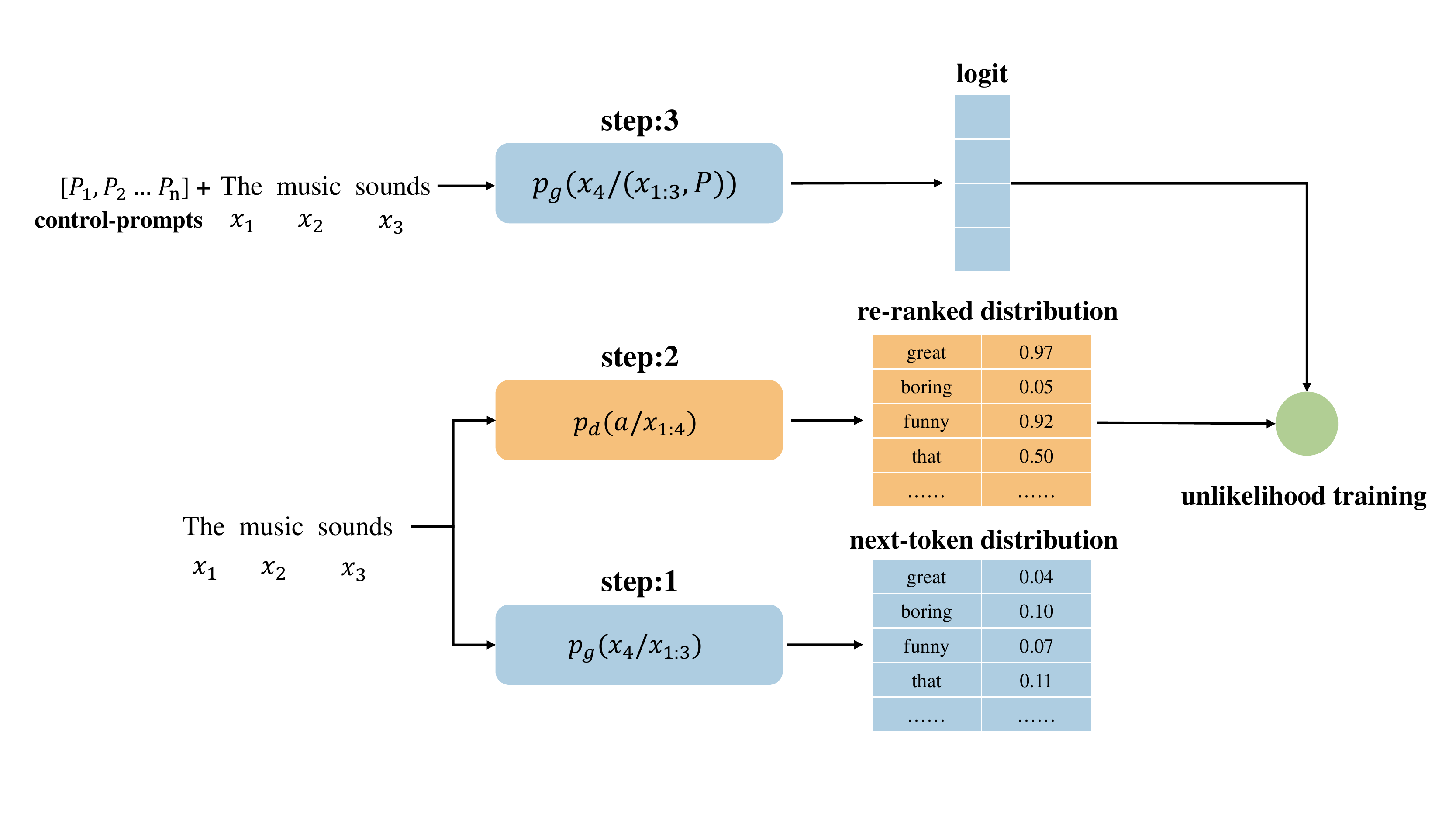} 
    \caption{An illustration of DisCup. The CLM  $p_{g}$ is always fixed. We first feed $x_{1:3}$ into the CLM model and get the next-token distribution (i.e, probable token distribution of $x_4$). Then, the \textit{attribute-discriminator} $p_{d}(a/x_{1:4})$ is employed to assign the probability classified as desired-attribute $a$ (e.g., positive sentiment) for each top-$k$ candidate token. Finally, we encourage $p_{g}(x_4/(x_{1:3},P))$ to generate tokens with a higher probability towards the desired attribute and avoid the ones with the opposite attribute, using \textit{unlikelihood training}.} 
    \label{approach}
\end{figure*}

\section{Methodology}
\subsection{Overview}


Different from fine-tuning the whole CLM model, we aim at optimizing some continuous virtual tokens (also called control-prompts) as a prefix, to steer a fixed CLM to generate attribute-specific texts. As illustrated in Figure\ref{approach}, we first feed a partial sequence into the fixed CLM to obtain a next-token prediction distribution. Then the attribute-discriminator is used to re-rank the top-$k$ candidates. Instead of using the traditional MLE objective (detailed in Appendix~\ref{mle_detail}), which maximizes the probability of the next token for a given context, we encourage the CLM within control-prompts to generate the candidate tokens classified by discriminator with higher confidence towards the desired attributes, and keep away from the opposite ones. The optimization weight of each candidate is determined by the normalized attribute confidence of the discriminator.

\subsection{Model Architecture}
Similar to the vanilla prompt tuning~\cite{tailor_2022}, we aim to get the continuous control-prompts $P_k$ for each attribution $A_k$. Let us denote the dimensionality of word embeddings  in the CLM as $d_{e m b}$, and the prompt length of  $A_k$ as $\ell_k$. The attribute control-prompts are represented as:
\begin{equation}
P_{k} \in \mathbb{R}^{\ell _{k} \times d_{e m b}},
\end{equation}
where the parameters are initialized randomly. Following the paradigm described in Equation~\ref{forward}, with the prompts $P_k$ as a prefix, the probability of modeling the text $X_n=\{x_1, x_2, \dots, x_n\}$ with a CLM  parameterized with $\theta$ can be formulated as:

\begin{equation}
\label{forward}
P_{\theta}\left(x_{1: n}\right)=\prod_{t=1}^{n} p_{\theta}\left(x_{t} \mid x_{<t}, P_k \right).
\end{equation}

Empirically,  we re-parameterize the $P_k$ for stable training. An external $LSTM_{\theta^\prime}$ module is introduced to make the control-prompts close to the natural language. Formally, it is given by:
\begin{equation}
P_{k}[i,:]=\operatorname{LSTM}_{\theta^\prime}\left(P_{k}^{\prime}[i,:]\right), 
\end{equation}
where $i \in [0, \ell_{k})$, and $\theta^\prime$ denotes the parameters of $LSTM$. During training phrase, we fix the LM's parameters $\theta$ and only update $\theta^\prime$.

\subsection{Candidate Token Selection with Attribute-discriminator}

Instead of using the context's next token in the training corpus as ground-truth tokens, we use an external attribute-discriminator to select candidates from the next-token distribution given by a frozen CLM. Since the CLM is trained with a variety of text corpora, extracting the candidate tokens from the fixed CLM can improve the generalization ability of the CTG model. Additionally, selecting the most highly probable tokens produced by the fixed CLM as ground-true tokens, ensuring the trained CTG model remains near the original CLM.

More specifically,  we first train an \textit{attribute-discriminator} $p_{d}(a/x)$ based on given labelled training data.  Then assuming a completed sample $X_n=\{x_1,x_2,\dots,x_n\}$,  we get the candidate tokens distribution via a CLM at arbitrary step $t$. Formally, we use $\mathbf{h}_{t} \in \mathbb{R}^{|\mathcal{V}|}$ to represent the output logit, and calculate the probability distribution over the vocabulary $\mathcal{V}$:

\begin{equation}
\label{logit}
p_{\theta}\left(\hat{\boldsymbol{x}_{t}} \mid x_{<t}\right)=\operatorname{softmax}\left(\mathbf{h}_{t}\right),
\end{equation}
where $\hat{\boldsymbol{x}_{t}}$ is a normalized distribution over the vocabulary, and a higher value of it indicates that the corresponding token is more likely to be combined with the context $x_{<t}$ into a fluent sentence. In order to maintain the fluency, we choose top-$k$ tokens as the \textbf{re-ranked candidate tokens} $\mathcal{C}$.  They are then concatenated with the context $x_{1:t-1}$ respectively, forming a partial sequence $\mathbf{X}_{1:t}^c$ that is fed into the discriminator to obtain the classification confidence of the desired control attribute $a$:

\begin{equation}
\mathbf{d}[c] = p_{d}\left(a \mid \mathbf{X}_{1: t}^c\right),
\end{equation}
where $ c \in \mathcal{C}$, every candidate token $c$ is scored by the attribute discriminator; the higher the probability, the closer the corresponding partial sentence is to the desired attribution. In reverse, we also calculate the unlikely probability of the candidate tokens for further unlikelihood training. And the re-ranked token's attribute probability can be redefined as:
\begin{equation}
 \mathbf{d}^{\prime}[c] = 1- \mathbf{d}[c].
\end{equation}

\subsection{Unlikelihood Training}
Inspired by the unlikelihood training used in previous work~\cite{Welleck2020Neural}, we expect the CTG model to generate the tokens recognized as the desired-attribute at higher confidence by the discriminator, and in contrast, keep away from the tokens with lower confidence. Therefore, the loss function is composed of two parts: \textit{likelihood objective} and \textit{unlikelihood objective}.

Specifically, we first use a $softmax$ function to normalize the candidate's probability distribution re-ranked by the discriminator: 
\begin{equation}
\boldsymbol{s}[c] = \operatorname{softmax}(\mathbf{d}[c] / \alpha),
\end{equation}
where $\alpha$ is the temperature used to control the sharpness of probability distribution. The smaller the temperature, the distribution is closer to its one-hot form. We conduct the same operation on $\mathbf{d}^{\prime}[c]$, and get $\boldsymbol{s^{\prime}}[c]$.  After that, given the trainable control-prompts, we concatenate it with $x_{<t}$, then feed it into the CLM model. The next token distribution $p_{\theta}\left(\hat{\boldsymbol{x}_{t}} \mid x_{<t}, P_k\right)$ can be obtained by forward propagation. On the side of likelihood optimization, we encourage the CLM model to generate the tokens with a higher probability scored by the discriminator toward the desired attribute. Therefore, the \textit{likelihood objective} is:
\begin{equation}
\mathcal{L}_{like}\left(x_{t}\right) = -\sum_{c \in \mathcal{C}} \boldsymbol{s}[c] \log p_{\theta}\left(c \mid x_{<t}, P_k\right).
\end{equation}

On the contrary,  \textit{unlikelihood objective} is used to keep the generated tokens of the CTG model away from those unlikely candidates, which can be intuitively formulated as:

\begin{equation}
\mathcal{L}_{unlike}\left(x_{t}\right) = -\sum_{c \in \mathcal{C}}\mathbf{s}^{\prime}[c] \log(1-p_{\theta}\left(c \mid x_{<t}, P_k\right).
\end{equation}
Finally, we use $x_{t}^{(i)}$ to represent the step $t$ of the $i$-th sample in the given training dataset $\mathcal{D}$, and the objective of discriminator cooperative unlikelihood prompt-tuning is defined as:

\begin{equation}
\label{mse}
\mathcal{L}\left(\theta, \mathcal{D}\right)=\sum_{i=1}^{|\mathcal{D}|} \sum_{t=1}^{\left|\mathbf{x}^{(i)}\right|}  \mathcal{L}_{like}\left(x_{t}^{(i)}\right)+ \mathcal{L}_{unlike}\left(x_{t}^{(i)}\right),
\end{equation}
We provide a simplified theoretical analysis from the gradient perspective in Appendix~\ref{theoretical}.

%% file: experiment.tex
\section{Experiments}

\begin{table*}[t]


\resizebox{\linewidth}{!}{
\begin{tabular}{llcccccc}
\toprule[1pt]
                                            &                                    & \multicolumn{3}{c}{\textbf{Correctness(\%)}($\uparrow$)}          & \multicolumn{1}{l}{\textbf{Fluency}($\downarrow$ )} & \textbf{Diverty}($\uparrow$)                         & \multicolumn{1}{l}{\textbf{Coverage}($\downarrow$)} \\
\multirow{-2}{*}{\textbf{Target Sentiment}} & \multirow{-2}{*}{\textbf{Method}}  & Positive    & Neutral & Negative    & PPL                   & Dist-1/Dist-2/Dist-3 & Rate(\%)                 \\ \midrule[1pt]

                                & $\text{PPLM}^{\heartsuit}$                            & \multicolumn{1}{l}{} & 52.68            & 8.72                 & 113.54          
                                            & 0.39/0.83/0.89                                              & 3.47                                  \\ 
                                            & $\text{DAPT}^{\spadesuit} $                             & \multicolumn{1}{l}{} & 61.81            & 14.17                & 41.89                                & 0.20/0.64/0.84                                              & 4.22                                \\
                                            & $\text{CTRL}^{\spadesuit} $                            & \multicolumn{1}{l}{} & 77.24            & 18.88                & 48.24                                & 0.13/0.53/0.79                                              & 8.86                                \\
                                            & $\text{GEDI}^{\heartsuit} $                               & \multicolumn{1}{l}{} & 86.01            & 26.80                 & 123.56                               & 0.20/0.66/0.85                                              & 3.12                                \\
                                            & $\text{DEXPERT}^{\heartsuit} $                            & \multicolumn{1}{l}{} & 94.46            & 36.42                 & 60.64                                & 0.18/0.63/0.84                                            & 3.49                                   \\
                                            \cline{2-8}
                                             & $\text{Vanilla Prompt-tuning}^{\spadesuit} $                    & \multicolumn{1}{l}{} & 78.08            & 40.88                & 38.23                       
                                            & 0.14/0.48/0.73                                              & \uline{69.60}            \\ 
                                            & \multicolumn{1}{l}{$\text{FUDGE (discriminator-based)}^{\heartsuit} $}  & \multicolumn{1}{l}{} & \textbf{96.92}   & 56.04                & \uline{228.76}                               
                                            & 0.16/0.52/0.76                                              & 1.78                                \\
                                            & $\text{DisCup}^{\spadesuit} (\textit{w/o unlikelihood}) $                         & \multicolumn{1}{l}{} & 91.58      & 49.92       & 44.20                       
                                            & 0.15/0.51/0.77                     & 3.94  \\
                                            
\multirow{-8}{*}{Positive}                  & $\text{DisCup}^{\spadesuit} (\textit{w/ unlikelihood}) $                             & \multicolumn{1}{l}{} & 94.98   & \textbf{64.96}       & 48.71                       
                                            & 0.14/0.50/0.76                     & 3.24                                \\ \hline
                                            & $\text{PPLM}^{\heartsuit} $                                & 10.26                & 60.95            & \multicolumn{1}{l}{} & 122.41                              & 0.40/0.83/0.90                                              & 3.47                               \\
                                            & $\text{DAPT}^{\spadesuit} $                              & 12.57                & 66.72            & \multicolumn{1}{l}{} & 43.01                                & 0.19/0.63/0.83                                              & 3.33                                \\
                                            & $\text{CTRL}^{\spadesuit} $                              & 20.95                & 62.37            & \multicolumn{1}{l}{} & 45.27                                & 0.13/0.51/0.78                                              & 9.87                                \\
                                            & $\text{GEDI}^{\heartsuit} $                               & 60.43                & 91.27            & \multicolumn{1}{l}{} & 138.93                               & 0.19/0.66/0.86                                              & 4.11                                \\
                                            & $\text{DEXPERT}^{\heartsuit}$                             & 64.01                & \textbf{96.23}   & \multicolumn{1}{l}{} & 67.12                                & 0.20/0.64/0.83                                              & 2.71                                \\ \cline{2-8} 
                                            & \multicolumn{1}{l}{$\text{Vanilla Prompt-tuning}^{\spadesuit}$} & 49.28                & 73.20            & \multicolumn{1}{l}{} & 39.55                                & 0.14/0.49/0.72                                              &  \uline{56.56}       \\
                                            & \multicolumn{1}{l}{$\text{FUDGE (discriminator-based)}^{\heartsuit} $ }  & 66.84                & 98.76   & \multicolumn{1}{l}{} & \uline{265.79}                               & 0.23/0.68/0.83                                              & 1.29                                \\
                                            
                                            & \multicolumn{1}{l}{$\text{DisCup}^{\spadesuit}(\textit{w/o unlikelihood}) $}     & 60.80       & 90.64         & \multicolumn{1}{l}{}  &36.72               
                                            & 0.12/0.45/0.72                     & 3.51  \\
\multirow{-8}{*}{Negative}         & $\text{DisCup}^{\spadesuit} (\textit{w/ unlikelihood}) $                             & \textbf{68.76}       & 93.64            & \multicolumn{1}{l}{} & 45.60    & 0.12/0.48/0.77                                              & 2.97                                \\ \bottomrule[1pt]
\end{tabular}
}
\caption{The main experimental results of sentiment controllable text generation.   $\uparrow$ indicates that the higher corresponding value is better, and $\downarrow$ is the opposite. ''—" represents the outlier of the items, and the corresponding methods are just regarded as the reference, but not included in the performance comparison. $\heartsuit$  and $\spadesuit$ mean the \textit{decoding-time} and the \textit{training} approaches respectively.}

\label{automatic_seniment_result}

\end{table*}

\subsection{Evaluation Metric}

\noindent \textbf{Automatic Evaluation}. We test the generated texts from three aspects.  (1)\textbf{ Attribute Relevance}: we use an external sentiment classifier provided by Huggingface\footnote{ \url{https://huggingface.co/distilbert-base-uncased-finetuned-sst-2-english} \label{sentiment_classfier}} to test whether the generated texts satisfy the controllable sentiment attribute, and count the proportion of samples that conform to target sentiment as a quantitative indicator, called \textbf{Correctness}. As for the toxicity avoidance task, we use the Perspective API\footnote{\url{https://perspectiveapi.com/} \label{toxic_api}} to calculate the \textbf{Average Toxicity Probability} for the generated texts. (2) \textbf{Fluency}: GPT2-large is used to calculate the Perplexity (PPL), which reflects the text's fluidity. (3)\textbf{Diversity}: Distinctness is employed to measure the text's Diversity. Specifically, this is done by calculating the numbers of uni-grams, bi-grams and tri-grams among all the generated texts and then counting their proportion in all words. The result is reported as Dist-1/2/3. Furthermore, we design a domain generalization metric for the CTG model. Specifically,  we collect some domain-specific words from the training corpus and calculate the proportion of sentences that contain the domain-specific words among all generated sentences. This metric is named \textbf{Coverage Rate}. The details of domain-specific words can be seen in Appendix~\ref{keyword_movie}.

\noindent \textbf{Human Evaluation}. We also conduct human evaluation from three aspects. \textbf{Relevance} reflects the degree of achievement for the desired control attribute. \textbf{Topicality} means whether the generated continuations are consistent with the given prompts. \textbf{Fluency} evaluates the text's fluency from the human perspective. Detailed information can be seen in Appendix~\ref{human_evaluation}.

\subsection{Baselines}
We empirically compare the proposed method with a wide range of baselines.

\noindent \textit{Training Approaches}: (1) Conditional Transformer LM  (\textbf{CTRL}~\cite{ctrl}) is a pre-trained language model conditioned on task-specific control codes. (2) Domain-adaptive pre-training (\textbf{DAPT}~\cite{gururangan-etal-2020-dont}) is an approach that applies the PLM to the domain of a target task. We  adapt it to sentiment control by pre-training it on sentiment corpus.

\noindent \textit{Decoding-time Approaches}: (1) \textbf{PPLM} is a typical decoding-time method, which uses the discriminator to update PLM's hidden layer to steer controllable generation~\cite{pplm}. (2) \textbf{GEDI}~\cite{gedi} fine-tunes external CCLMs as generative discriminator to guide the attribute-specific generation. (3) \textbf{DEXPERT}~\cite{liu-etal-2021-dexperts} is the state-of-the-art (SOTA) model so far, which use fine-tuned GPT2 as an expert/anti-expert to steer the text generation, and we directly choose GPT2-large as guided module. For the above baselines, we use GPT2-large as the base CLM, and the detailed settings are consistent with the existing work~\cite{liu-etal-2021-dexperts} \footnote{code is available at \url{https://github.com/alisawuffles/DExperts} \label{dexpert_code}}.

\noindent \textit{Discriminator-based}: We also explore a decode-time reference baseline, which is closely related to our proposed method. The principle is the same as \textbf{FUDGE}~\cite{yang-klein-2021-fudge}, and the generative model is GPT2-large. The difference is that we replace the \textit{future discriminator} with the \textit{attribute-discriminator} trained based on GPT2-small. In order to provide more reference information,  we keep the depth of top-k sampling ($k=70$) at the same level as the size of re-ranked candidate tokens $\mathcal{C}$ in our method. 

\noindent \textit{Vanilla Prompt-tuning}: It is a base version of DisCup,  and each attribute control-prompt is trained under the attribute-specific corpus (e.g., positive and negative sentiment), which is similar to \textbf{Tailor}~\cite{tailor_2022}. GPT2-large is used as the base CLM model and its parameters and sampling algorithm used in the experiments are consistent with our method.

\subsection{Sentiment Control Task}
\subsubsection{Experimental Setup}
\textbf{Dataset.} Following the previous work~\cite{liu-etal-2021-dexperts},  we take the widely used Stanford Sentiment Tree (SST-5)~\cite{socher-etal-2013-recursive}  as the training corpus, which is collected from movie reviews. The generation prompts come from different domains collected from OpenWebText (more details can be seen in ~\cite{liu-etal-2021-dexperts}). In total, we get 5K neutral prompts, 2.5K negative prompts, and 2.5k positive prompts, and the affective polarity of those prompts was measured based on the natural generation of GPT2-large. During the experiments, we uniformly set the maximum length of the generation to 20 tokens for every prompt.

\noindent \textbf{Method Setting.} we fine-tune GPT2-small on SST-5 as the expert of attribute classifier and use GPT2-large as the base CLM. We conduct the discriminator cooperative unlikelihood prompt tuning on SST-5, ignoring the text sentiment label. The detailed settings of other hyper-parameters can be seen in Appendix~\ref{hyper_parameter}.

\subsubsection{Results and Analysis}

\noindent \textbf{Automatic Evaluation Result.}  As shown in Table~\ref{automatic_seniment_result}, our method significantly outperforms most baselines in correctness and text fluency. PPLM is the least effective because updating the PLM's hidden layer destroys the structure of the original PLM and results in arbitrary output. Compared with the training methods (i.e., DAPT, CTRL), decoding-time methods (i.e., DEXPERT, GEDI) show better controllability yet lower text fluency, which suggests decoupling the guided model from generative models will hurt the quality of the generated text. DEXPERT's performance is close to our method. However, our method shows a higher degree of symmetry in the adversarial steering (i.e., negative prompts toward positive continuations, and vice versa.). In particular, the correctness of our method outperforms DEXPERT by 28.54\% in steering the negative prompts to positive generation. We suspect that using a fine-tuned CLM as an expert may itself be biased, and the discriminator can relieve this problem well. DisCup and the vanilla prompt learning both show slight drops in terms of diversity. Fortunately, we find that there is a trade-off between diversity and PPL, and with the increase in the depth of sampling, this gap narrows. More details can be seen in Appendix~\ref{diversity_ppl}.

The discriminator-based method shows a potential to achieve a good control ability without considering the degree of text fluency. On the contrary, vanilla prompt-tuning could produce fluent texts yet shows a poor performance both in correctness and domain generalization. Specifically, around 60\% of the generated texts contain domain-related keywords from the training corpus. \textbf{The concrete examples can be seen in Table~\ref{vanilla_example}}. Our method inherits the advantages of both prompt-tuning and discriminator-based approaches, thus does not suffer from domain generalization issues, with only less than 4\% of generated sentences containing the domain-related keywords. Even without unlikelihood training, our method still far outperforms vanilla prompt-tuning, since the candidates are re-ranked by the discriminator and incorporate inter-attribute relationships. The unlikelihood training further improves the correctness, $\sim$ 3\% for the steering from neutral prompts to positive/negative, and $\sim$10\% for the adversarial steering.

\noindent \textbf{Human Evaluation Result.}  As shown in Table~\ref{human_evaluate}, the human evaluation results are almost consistent with the automatic evaluation. DisCup outperforms the baselines. However, the vanilla prompt-tuning has lower scores in Fluency and Topicality. The reason is that vanilla prompts usually steer GPT2 to generate texts about movies (seen specific examples in Table~\ref{vanilla_example}), making human evaluators feel like it is always digressing. Our method can alleviate this problem, as our optimization goal is to predict candidate tokens naturally self-generated by the frozen CLM, thus keeping the learned control-prompts not steer too far away from the original language model. 

\begin{table}[t]
\resizebox{\linewidth}{!}{

\begin{tabular}{lccc}
\toprule[1pt]
\multicolumn{1}{l}{\textbf{Method}} & \textbf{Relevance($\uparrow$)} & \textbf{Fluency($\uparrow$)} & \textbf{Topicality($\uparrow$)} \\ 
\midrule[1pt]
$\text{CTRL}^{\spadesuit}$                                & 4.7                & 6.4               & 6.7                 \\
$\text{Vanilla Prompt-tuning}^{\spadesuit}$               & 5.3                & 6.4               & 6.3                 \\
$\text{DEXPERT}^{\heartsuit} $                            & 5.8                & 6.5               & 6.9                 \\
$\text{DisCup}^{\spadesuit}  $                           & \textbf{7.8}       & \textbf{6.9}      & \textbf{7.1}        \\ 
\bottomrule[1pt]
\end{tabular}
}
\caption{The human evaluation results on sentiment  control experiment.}
\label{human_evaluate}
\end{table}

\begin{table}[htbp]
\centering
\renewcommand\arraystretch{0.93} 

\resizebox{\linewidth}{!}{
\begin{tabular}{lccc}
\toprule[1pt]
\multicolumn{1}{l}{\multirow{2}{*}{\textbf{Method}}} & \textbf{Toxicity($\downarrow$)} & \textbf{Fluency($\downarrow$)} & \textbf{Diversity($\uparrow$)}   \\
\multicolumn{1}{c}{}                                 & Avg. toxicity prob & PPL              & dist-1/dist-2/dist-3 \\ \midrule[1pt]
$\text{PPLM}^{\heartsuit}$                                                 & 0.121             & 48.02            & 0.33/0.79/0.91             \\
$\text{GEDI}^{\heartsuit}$                                                 & 0.091             & 56.94            & 0.18/0.66/0.87             \\
$\text{DAPT}^{\spadesuit}$                                                 & 0.089             & 47.03            & 0.17/0.62/0.85             \\
$\text{DEXPERT}^{\heartsuit}$                                              & 0.086             & 49.71            & 0.18/0.64/0.86             \\ \hline
$\text{FUDGE(discriminator-based)}^{\heartsuit}$                          & \textbf{0.062}    & \uline{354.78}   & 0.20/0.67/0.83             \\
$\text{Vanilla Prompt-tuning}^{\spadesuit}$                                       & 0.108                & 27.40            & 0.12/0.47/0.74             \\

$\text{DisCup}^{\spadesuit} (\textit{w/o unlikelihood})$                                                & 0.066                & 39.84            & 0.18/0.60/0.83             \\ 
$\text{DisCup}^{\spadesuit} (\textit{w/ unlikelihood})$                                                 & 0.064              & 39.82            & 0.17/0.62/0.84             \\ \bottomrule[1pt]
\end{tabular}
}
\caption{The main experimental results on toxicity avoidance. Underscores indicate outliers, and the corresponding method is just regarded as the reference, but not included in the performance comparison. $\heartsuit$ and $\spadesuit$ represent the \textit{decoding-time} and the \textit{training} approaches respectively.}

\label{detoxic}
\end{table}

\subsection{Toxicity Avoidance Task}
\subsubsection{Experiment Setting}
\textbf{Dataset.} Toxicity training data is provided by Jigsaw Unintended Bias in Toxicity Classification Kaggle challenge \footnote{\url{https://www.kaggle.com/c/jigsaw-unintended-bias-in-toxicity-classification} }. The dataset contains around 160K toxic comments and 1.4M nontoxic comments. As for the generation prompts, we follow the previous work~\cite{liu-etal-2021-dexperts} and use 10K nontoxic prompts from the RealToxicityPrompt~\cite{gehman-etal-2020-realtoxicityprompts}.

\noindent \textbf{Method Setting.} Following the setting in the sentiment control task, we fine-tune GPT2-small as an attribute classifier on the toxicity dataset, and use GPT2-large as the base CLM. During the prompt tuning phrase, we randomly sample 5K toxic and 5K nontoxic, respectively, from the toxicity dataset as a training corpus. The whole nontoxic data in the dataset is used for vanilla prompt learning, ensuring the
prompts learn the features of non-toxic data. The more detailed settings are given in Appendix~\ref{hyper_parameter}.

\subsubsection{Result and Analysis}
As shown in Table~\ref{detoxic}, the average toxicity probability of DisCup is 0.064, which significantly outperforms 0.086 achieved by DEXPERT, a SOTA model so far. The performance of DisCup is slightly lower than FUDGE, a reference baseline with lower text quality (the PPL is over 350). The vanilla prompt-tuning performs poorly, which is expected due to its inability to learn from toxic texts. As for the PPL, the vanilla prompt-tuning fits nontoxic text without harming the text quality of PLM, and thus it shows a better performance. DisCup is slightly inferior to it but better than other baselines, even though we set the same size of candidate tokens as that in top-k decoding algorithm of FUDGE. Because control-prompts are co-training with PLMs, more interaction would allow the model to maintain the original characteristics of CLM as much as possible.

Different from the sentiment control, DisCup remains competitive in term of text diversity, which is close to GEDI, DAPT, and DEXPERT. The main reason is that the range of nontoxic text is relatively more expansive than the toxic, thus allowing the learned control-prompts to imitate the trait and generate diverse texts. Similar to sentiment control, the branch without unlikelihood training is inferior to the setting with an unlikelihood objective, further indicating that the unlikelihood training 
can provide a gain of control performance.

\begin{figure}[t] 
    \centering 
    \includegraphics[width=0.4\textwidth]{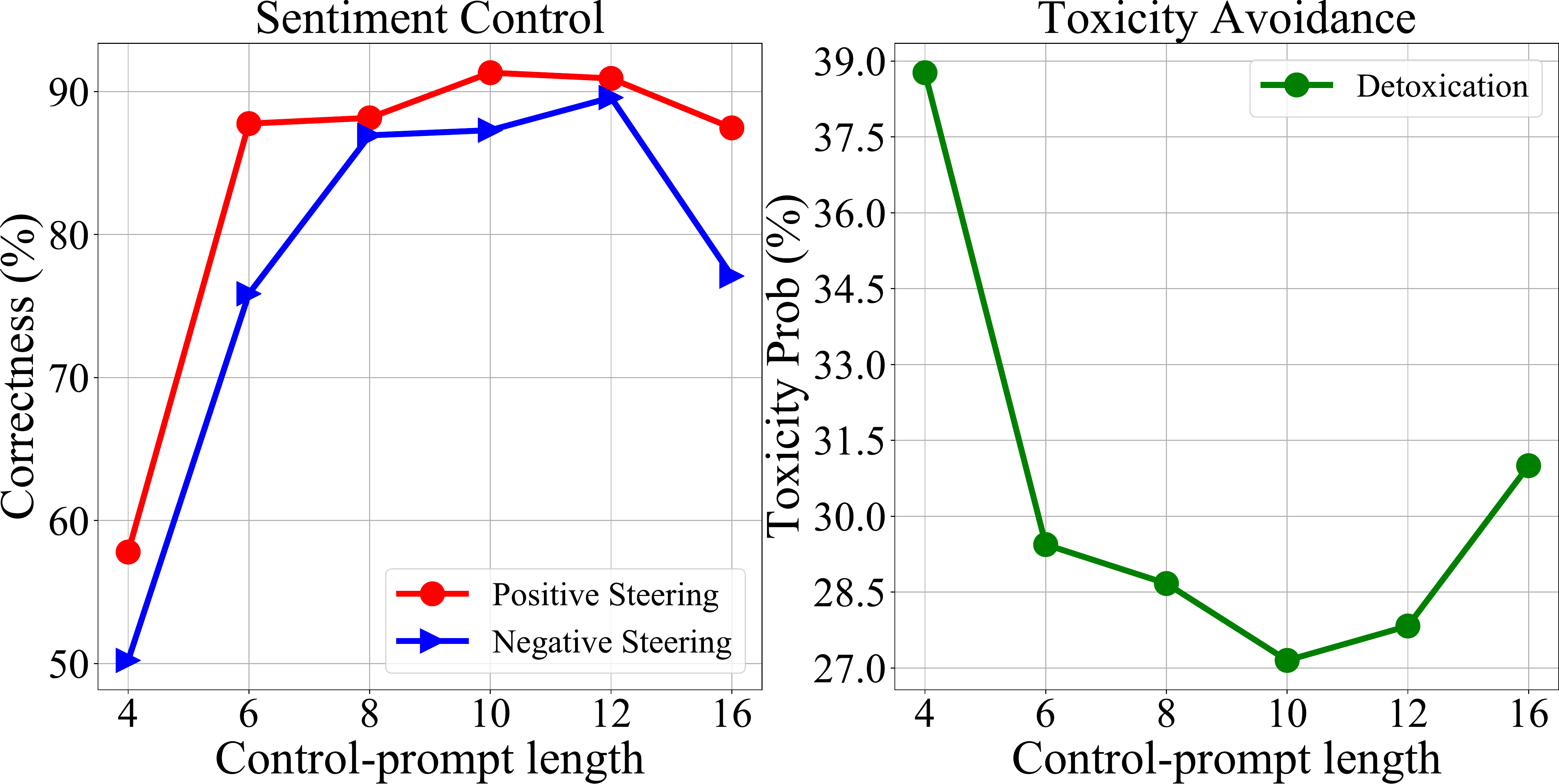} 
    \caption{The effect of control-prompt length on model performance. Toxicity probability is measured under an offline  classifier  trained based on GPT2-small. The size of the re-ranked candidates is set to 30 on the both tasks.} 
    \label{prompt_length}
\end{figure}

\subsection{Further Analysis}

\begin{table}[]
\centering
\begin{tabular}{lc}
\toprule[1pt]
\multicolumn{1}{l}{\textbf{Method}} & \textbf{Time Cost(second)} \\ \midrule[1pt]
PPLM                                & 37.39                            \\
DEXPERT                             & 2.54                             \\
GEDI                                & 1.86                             \\
GPT2-large                          & 0.78                             \\
DisCup                                & 0.94                             \\ \bottomrule[1pt]
\end{tabular}

\caption{The cost (time) for generating 20 tokens.}
\label{inference_speed}
\end{table}

\noindent \textbf{Control-prompt Length.} We investigate the effect of control-prompt length on the control performance. The result is shown in Figure~\ref{prompt_length}, which suggests that around 10 continuous tokens are enough to achieve a competitive performance, while too long control-prompts can cause difficulty in optimization and hurt the performance. This highlights the promise of prompt learning in parameter-efficient CTG.

\textbf{Inference Speed.} The relatively short control-prompt allows our method to be equipped with efficient inference speed. As shown in Table~\ref{inference_speed}, our method outperforms all decode-time approaches, with an inference speed closer to the pure PLM (GPT2-large). Specifically, our method generates 20 tokens in only 0.94 seconds (that is 0.78s for GPT2-large). 


\begin{figure}[h] 
    \centering 
    \includegraphics[width=0.4\textwidth]{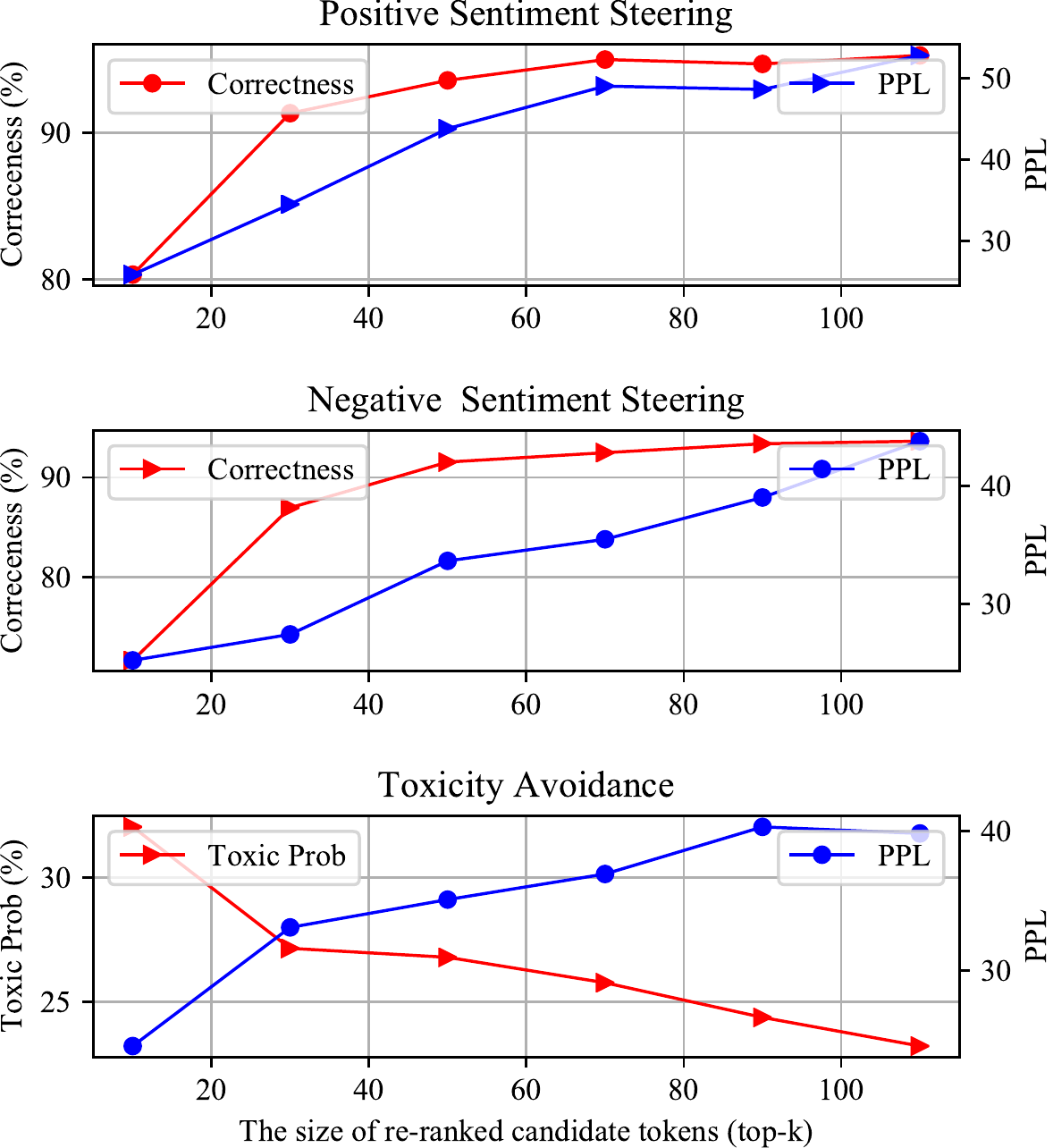} 
    \caption{The effect of size of re-ranked candidate tokens on the model control performance and text fluency. The prompt length is set to 10 for positive sentiment steering and toxicity avoidance and 12 for negative sentiment steering. The toxicity probability is measured with an offline classifier based on GPT2-small.
    } 
    \label{top_k}
\end{figure}


\textbf{The size of candidate tokens $\mathcal{C}$.}  The control performance of our method is proportional to the size of candidate tokens $\mathcal{C}$. As shown in Figure~\ref{top_k}, the deeper the sampling scope is, the better the control performance will be, but meanwhile the PPL deteriorates. This phenomenon is intuitive in a sense that the wider the candidate tokens, the more likely it is to provide tokens satisfying the control conditions. However, there will be more selected tokens in low-probability regions, resulting in a decrease in text fluency. 
\begin{figure}[t] 
    \centering 
    \includegraphics[width=0.30\textwidth]{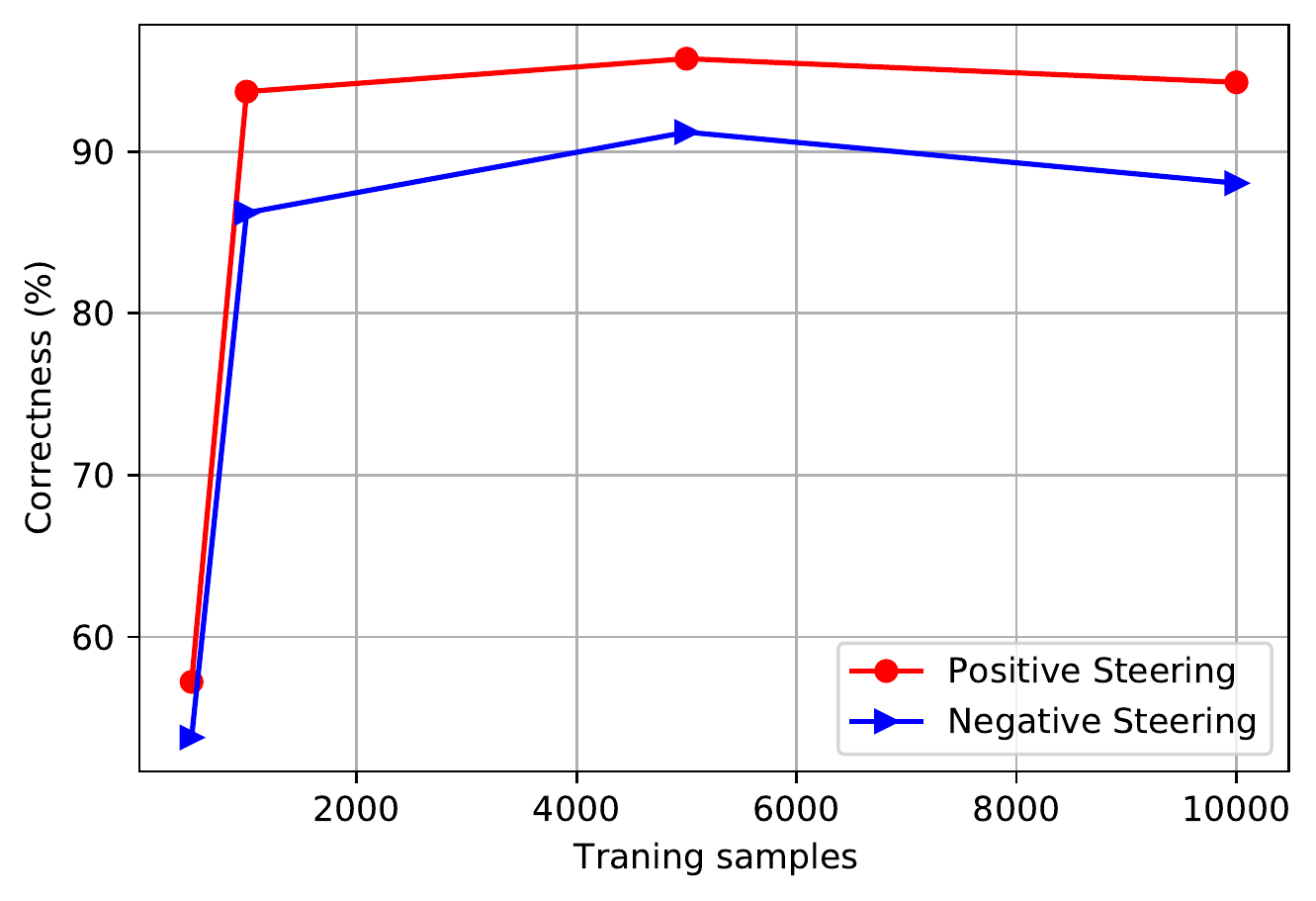} 
    \caption{The effect of training samples on sentiment control performance. The size of the re-ranked candidates is set to 70, prompt length is 10 on the both tasks. The training samples are randomly selected from OpenWebTextCorpus~\cite{Gokaslan2019OpenWeb}.} 
    \label{sample_size}
\end{figure}

\textbf{The size of training samples.} We have also conducted an additional analysis to explore the size of training samples on sentiment control performance. We assume that the discriminator is already trained, and DisCup is then trained on some unlabeled samples. As shown in Figure~\ref{sample_size}, the size of the training data is essential to the result, and 1K samples are enough to get a competitive result, more than 5K samples could achieve the optimal result.

%% file: conclusion.tex
\section{Conclusions}
We have proposed a novel alternative for attribute-controllable text generation, namely \textit{discriminator cooperative unlikelihood prompt-tuning (DisCup)}. Instead of the traditional next-token prediction based on a training corpus, we use a attribute-discriminator to select unlikely/likely candidates from the tokens naturally self-generated by a frozen CLM, which is trained on multitudinous textual corpora and capable of producing texts with a high degree of diversity. Then, unlikelihood training is employed to optimize the control-prompts. Experiments conducted on two typical CTG tasks, i.e., sentiment control and toxicity avoidance, prove that our approach not only significantly outperforms the vanilla prompt-tuning approaches, but also exhibits superiority over the existing training and decoding-time approaches.

%% file: limitation.tex
\section*{Limitations}
DisCup collects (un)likely candidate tokens from an original pre-trained CLM, instead of through next-token prediction based on the training corpus. This means that the performance of our approach will be immensely correlated with the power of the base CLM. If the base language model is of a poor quality, the training procedure trends to guide the CTG model to produce awkward tokens provided by the CLM. Consequently the learned control-prompts will steer the CLM to generate poor-quality texts at the inference stage.
 
 In addition,  DisCup has been shown to achieve a competitive performance in terms of attribute controllability, text quality, number of model parameters, and inference speed. However, this comprehensive ability is limited in the attribute control task so far, and it is hard to directly apply our method to fine-grained controlled text generation scenarios such as Table-to-Text. This is an open problem that we will explore in our future work.

\section*{Acknowledgments}
This research was supported in part by Natural Science Foundation of Beijing (grant number:  4222036) and Huawei Technologies (grant number: TC20201228005).

%% file: appendices.tex
\section*{Appendices}

\appendix
In this section, we provide external information for reproducing the experimental results mentioned in the paper and some additional preliminaries, theoretical analysis, and experimental results supplementary to the main page.

\section{Preliminaries}
\subsection{Language Modeling}
\label{language_model}

\textbf{Casual Lanauge Model(CLM)} usually leverage a auto-regressive PLM to perform  step-wise density estimation(i.e., next-token prediction). Suppose a CLM  parameterized with $\theta$, given the partial sequence $x_{<t}$, it assigns a probability  $p_{\theta}(x_{t}| x_{<t})$ for every token over a vocabulary $\mathcal{V}$ at next-token $x_t$ generation. When generating a sequence text $X_{n}=\{x_1, x_2, \dots, x_n \}$,  it could be formulated by chains rules as follow:

\begin{equation}
\label{forward}
P_{\theta}\left(X_{n}\right)=\prod_{t=1}^{n} p_{\theta}\left(x_{t} \mid x_{<t}\right).
\end{equation}
The whole process is conducted iteratively. Firstly, sampling a token at every step from $p_{\theta}\left(x_{t} \mid x_{<t}\right)$, and then the selected token  is concatenated with inputs for next step generation.

\subsection{Model Training with MLE} 
\label{mle_detail}
CLM is usually trained and accomplished with Maximum Likelihood Estimation (MLE). Given a finite set of training samples $\mathcal{D}$ and a CLM with parameters $\theta$, the optimized object is defined as follows:
\begin{equation}
\label{mse}
\mathcal{L}\left(\theta, \mathcal{D}\right)=-\sum_{i=1}^{|\mathcal{D}|} \sum_{t=1}^{\left|\mathbf{x}^{(i)}\right|} \log p_{\theta}\left(x_{t}^{(i)} \mid x_{<t}^{(i)}\right) ,
\end{equation}
where $|\mathcal{D}|$,  $|\mathbf{x}^{(i)}|$ represents the number of samples in dataset and the length of a sample sequence respectively. $x_{t}^{(i)}$ is the next-token of $x_{<t}^{(i)}$, and $x_{1:t}^{(i)}$ is a partial sequence truncated from training sample. The parameters $\theta$ is updated by maximizing the log likelihood (i.e., probability of next-token prediction).

\section{Theoretical Analysis }
\label{theoretical}
In this section, we give a brief theoretical analysis from the gradient perspective, referring to previous work\cite{Welleck2020Neural}. For a more intuitive understanding of the effect of the unlikelihood objective, we simplify our loss function proposed in DisCup, assuming that there is only a single likely and unlikely candidate at every step. 

 Assume a CLM with a vocabulary $\mathcal{V}$, the next-token prediction at step $t$ is formulated as $p=p_{\theta}(x_{t}^*| x_{<t})$, and $p \in \mathbb{R}^{|\mathcal{V}|}$ is the output of the logit $h$, given by a $\text{softmax}$ activation function. We use $p_i= softmax(h_i)$ to represent the probability of the i-th token in $\mathcal{V}$, estimated by the CLM. Assume that the $like$ and $unlike$ are the index of unlikely and likely token over the whole vocabulary, and the loss function in our approach at a single step is simplified  as:
\begin{equation}
\mathcal{L}_t = -\log p_{\text{like}} - \log (1-p_{\text{unlike}}).
\end{equation}

The gradient with respect to the logit $h_i$ could be calculated using the chain rule. Formally, it is defined as follows:

\begin{equation}
\begin{aligned}
\frac{\partial \mathcal{L}_t}{\partial h_{i}}
&=\frac{\partial \mathcal{L}_t}{\partial p_{i}} \frac{\partial p_{i}}{\partial h_{i}}
=\left(\mathbb{I}\left[i= \text{like}\right]-p_{i}\right) \\
& -\frac{p_{\text{unlike}}}{1-p_{\text {unlike}}}\left(\mathbb{I}\left[i={\text {unlike }}\right]-p_{i}\right).
\end{aligned}
\end{equation}

As for the logit $h_{like}$, the gradient with respect to it could be formulated as follow:
\begin{equation}
\frac{\partial \mathcal{L}_t}{\partial h_{\text{like}}} = 1-p_{\text{like}}\left(1-\frac{p_{\text {unlike}}}{1-p_{\text {unlike}}}\right).
\end{equation}

In the same way, the gradient of logit $h_{unlike}$ could be represented as:

\begin{equation}
\begin{aligned}
\frac{\partial \mathcal{L}_t}{\partial h_{\text{unlike}}} 
& =\left(0-p_{\text {unlike}}\right)- \frac{p_{\text {unlike}}}{1-p_{\text {unlike }}}\left(1-p_{\text {unlike}}\right) \\
&=-2*p_{\text {unlike}}
\end{aligned}
\end{equation}

Finally, for the tokens with the index $k$ in the vocabulary, where  $k != like$ and  $k != unlike$,  the gradient of the logit $h_k$ could be represented as:
\begin{equation}
\frac{\partial \mathcal{L}_t}{\partial h_{k}}  =
p_{k}\left(1-\frac{p_{\text {unlike}}}{1-p_{\text {unlike }}}\right).
\end{equation}

By looking at their gradient expressions, we know that the gradient of $h_{\text {like}}$ is always a positive value, and that of $h_{\text{unlike}}$ is a negative value. Therefore, model optimization is always in the direction of increasing the probability of the likely token and decreasing the probability of the unlikely token. As for other tokens that neither belong to likely nor unlikely tokens, as the value of $p_{\text {unlike}}$ increases, their gradients trend to increase from negative to positive with a boundary of 0.5.

\section{Experimental Details}
\label{hyper_parameter}
\textbf{Baselines.} We conduct the same control tasks and use the same datasets as the previous work~\cite{liu-etal-2021-dexperts}, and thus for the baselines including PPLM, DART, CTRL, GEDI and DEXPERT, we directly use the hyper-parameters, checkpoints, sampling algorithms, generated texts, and experimental results provided by the open source resources of ~\citet{liu-etal-2021-dexperts}. All above baselines are supervised on the attribute-specific corpus, and cover almost all the existing typical CTG approaches.

\noindent  \textbf{Training details.}  All the experiments presented in our paper are conducted on a single NVIDIA A6000 GPU. We implement our methods, vanilla prompt tuning, and discriminator-based method (FUDGE) with the Pytorch deeping learning framework and HuggingFace Transformers package. During the training stage, the optimizer is Adam with a learning rate of 1e-3. For our approach, we search the temperature $\alpha$ over the value $\{0.1, 0.01, 0.005, 0.001\}$, and finally chose $\alpha = 0.005$ for positive sentiment control,  $\alpha = 0.01$ for negative sentiment and detoxication. The control-prompt length is set to 10 for sentiment control and toxicity avoidance,  and 12 for negative sentiment steering. As for the size of re-ranked candidate tokens $\mathcal{C}$, we search the top-k values over $\{10,30,50,70,90,110\}$, and choose top-$k =70$ for positive sentiment control and top-$k =110$ for the task of negative sentiment control and toxicity avoidance. Every CTG model is trained with 6 epochs, and we choose the checkpoint with the best control performance. For the vanilla prompt-tuning, the control-prompt length is set to 10, and all other settings remain the same as DisCup.

\noindent  \textbf{Generation settings.} During the decoding phase, we apply the top-10 sampling algorithm for our approach and vanilla prompt tuning. As for the discriminator-based baseline (FUDGE), we apply the top-$70$ sampling algorithm, which is roughly consistent with the size of re-ranked candidate tokens used in DisCup.

\section{Diversity \&  PPL}
\label{diversity_ppl}

We observe that the text diversity and text fluency quantized by PPL are a trade-off.  As shown in Table~\ref{div_ppl}, when we increase the depth of token sampling (i.e., increasing the size of top-k), the diversity of generated texts will increase, yet with the burden of decreasing text fluency.
\begin{table}[h]
\centering

\renewcommand\arraystretch{1.2}
\resizebox{\linewidth}{!}{
\begin{tabular}{clcc}
\toprule[1pt]
\textbf{Top-k} & \multicolumn{1}{c}{\textbf{Dist1/2/3}} & \textbf{PPL} & \textbf{Correctness(\%)}\\  \midrule[1pt]
5             & 0.14/0.47/0.72                         & 38.6         & 95.30\\
10            & 0.14/0.51/0/77                         & 48.0         & 94.98\\
30            & 0.15/0.55/0.82                         & 76.6         & 93.56\\
50            & 0.16/0.57/0.84                         & 92.5         & 92.71\\ \bottomrule[1pt]
\end{tabular}}
\caption{The relationship between diversity and text fluency. The result is test under 5K neutral prompts, steering direction of sentiment control generation is positive.  With increasing of sampling depth, i.e., size of top-$k$, text diversity increases, and text fluency deteriorates.}
\label{div_ppl}
\end{table}




\section{Human Evaluation}
\label{human_evaluation}

We conduct the human evaluation experiment for the sentiment control task. During the experiment,  we chose 10 positive-steering prompts composed of neutral and positive prompts, 10 negative-steering prompts composed of neutral and negative prompts. Each prompt is tested under four representative CTG models (i.e., CTRL, DEXPERT, Vanilla prompt tuning, and our approach), and we totally get 80 samples. To this end, we invite 5 experts who are well-educated in English to score the samples from the aspects including sentiment relevance, topicality, and text fluency. Every human expert is asked to give a score in the range of 0-10 from those three perspectives for each sample; the higher the score, the better the text's quality. Every expert takes around 40 minutes to finish the evaluation test, and we calculate the average score of each metric for every CTG method under comparison, recording the final results in Table~\ref{human_evaluate}.

\section{Domain-related Keywords}
\label{keyword_movie}
We count the high-frequency words appeared in SST-5 dataset, and then manually screen 10 keywords related to the domain of movie review. Those keywords are: \textbf{``movie'', ``movies'', ``films'', ``film'', ``story'', ``director'' ,``directors'' ,``comedy'' , ``audience'',  ``drama''}.


\section{Generation Examples}
In this section, we provide some specific examples for different CTG methods, which could be seen in Table~\ref{vanilla_example}, Table~\ref{detoxic_example}, and Table~\ref{sentiment_example}.

\begin{table*}[]
\renewcommand\arraystretch{1.4} 

\resizebox{\linewidth}{!}{

\begin{tabular}{cll}
\hline
\textbf{Target sentiment} & \multicolumn{1}{c}{\textbf{Vanilla Prompt Tuning}}                                                                                                                                            & \multicolumn{1}{c}{\textbf{DisCup (Ours)}}                                                                                                                                                             \\ \hline
positive                  & {\color[HTML]{000000} \begin{tabular}[c]{@{}l@{}}In fact the team was on its way to a \$5 million\\  budget in only two days.A {\color{red}{movie}} like this could\\  easily get lost in\end{tabular}}     & \begin{tabular}[c]{@{}l@{}}In fact the team was on its way to winning,"\\  said the former president of the United States \\ While Obama was president and a Democrat, the\end{tabular}       \\ \hline
positive                  & \begin{tabular}[c]{@{}l@{}}The dismissive look on her face sold it, \\ and it shows... The result is an unassuming\\  yet charming {\color{red}{movie}}, one that will appeal to\end{tabular}               & \begin{tabular}[c]{@{}l@{}}The dismissive look on her face sold it, but it \\ made her smile again. "Thanks very much, \\ you are awesome!" "Thanks too!" she\end{tabular}                    \\ \hline
positive                  & \begin{tabular}[c]{@{}l@{}}But they are spread across a huge land area so \\ that the {\color{red}{audience}} has a lot more time to enjoy\\  the {\color{red}{film}}. It's not as if it '\end{tabular}                    & \begin{tabular}[c]{@{}l@{}}But they are spread across a huge land area \\ so you can enjoy a good day with good friends,\\  and they can be fun but you have to keep a\end{tabular}           \\ \hline
positive                  & \begin{tabular}[c]{@{}l@{}}Doctors there treat underserved Medicaid patients, \\ a segment that's not well represented elsewhere\\  in the {\color{red}{film}}. But that's just the beginning\end{tabular}  & \begin{tabular}[c]{@{}l@{}}Doctors there treat underserved Medicaid \\ patients but there are many others that have helped,\\ " he added. "We want to be part of their success\end{tabular}   \\ \hline
positive                  & \begin{tabular}[c]{@{}l@{}}There is a plausible sense that the {\color{red}{film}} could easily\\  have been a lot better if its {\color{red}{director}} had spent \\ more time with the characters\end{tabular}           & \begin{tabular}[c]{@{}l@{}}There is a plausible reason why he wants to help us. \\ It was one of the most amazing things.\end{tabular}                                                        \\ \hline
positive                  & \begin{tabular}[c]{@{}l@{}}It was a hot alpha release, the kind of {\color{red}{film}} to \\ watch for an {\color{red}{audience}} that might be interested\\  in the genre but not necessarily in the\end{tabular}         & \begin{tabular}[c]{@{}l@{}}It was a hot alpha release, so please don't hesitate \\ but enjoy :) Thanks for the great work\end{tabular}                                                        \\ \hline
positive                  & \begin{tabular}[c]{@{}l@{}}“I would recommend all U.S. {\color{red}{audiences}} to make\\  the  journey to see the {\color{red}{story}}... It's well worth the trip\end{tabular}                                           & \begin{tabular}[c]{@{}l@{}}“I would recommend all U.S. citizens to join and \\ support your efforts," said a U.S. citizen, \\ who is now working in\end{tabular}                              \\ \hline
negative                  & \begin{tabular}[c]{@{}l@{}}nominee questions key Obama-era rules of \\ engagement. The {\color{red}{film}} is too long and repetitive, \\ its plot too ponderous and its politics too confused\end{tabular} & \begin{tabular}[c]{@{}l@{}}nominee questions key Obama-era rules and rules \\ of federal court: not true no one could have been\\  prosecuted even for doing nothing \end{tabular} \\ \hline
negative                  & \begin{tabular}[c]{@{}l@{}}All I can say is that if this were an independent film, \\ it's a shame that the {\color{red}{director}} and {\color{red}{producers}} did\\  so little to\end{tabular}                          & \begin{tabular}[c]{@{}l@{}}All I can say is it sucks but if there ever are bad days, \\ I'll never ever forget the first one\end{tabular}                                                     \\ \hline
negative                  & \begin{tabular}[c]{@{}l@{}}Bateson and her team watched the same old {\color{red}{story}} \\ and didn 't realize it's a {\color{red}{movie}} until a couple of \\ scenes in the end\end{tabular}                           & \begin{tabular}[c]{@{}l@{}}Bateson and her team watched the {\color{red}{movie}} over and \\ over until a single frame didn't match the other one.\\  But not only did the scene\end{tabular}                \\ \hline
negative                  & \begin{tabular}[c]{@{}l@{}}Toronto: Between the Lines feels like the first {\color{red}{film}}\\  {\color{red}{directed}} by a young, inexperienced {\color{red}{actor}}.\\ I would have liked to\end{tabular}                            & \begin{tabular}[c]{@{}l@{}}Toronto: Between the bad weather and bad luck\\  (or bad luck) the team never played well and had\\  little time or chance\end{tabular}                            \\ \hline
negative                  & \begin{tabular}[c]{@{}l@{}}Stuff like that, I'd like to see the  {\color{red}{director's}}  head \\ explode. '.. The {\color{red}{film}} does have a good deal\end{tabular}                                                  & \begin{tabular}[c]{@{}l@{}}Stuff like that, I could write more but no matter\\  where I went there's not even one restaurant. \\ No one will\end{tabular}                                     \\ \hline
negative                  & \begin{tabular}[c]{@{}l@{}}And now, to prove that point, they're making a\\  {\color{red}{comedy}} of the year...  and it's not funny... and\end{tabular}                                                   & \begin{tabular}[c]{@{}l@{}}And now, to prove that point, no matter what, \\ it would be wrong and  pointless to try to do \\ something without anyone's consent."\end{tabular}                \\ \hline
negative                  & \begin{tabular}[c]{@{}l@{}}Visit GoldAnswers for help moving to\\  Office Space. The Big Lebowski (1998)\\ A {\color{red}{comedy}} about two friends who are\end{tabular}                                   & \begin{tabular}[c]{@{}l@{}}Visit GoldAnswers for help moving to Office 365\\  for less. Not only cannot you move without the proper \\ documentation, but it also won't be easy\end{tabular}  \\ \hline
\end{tabular}
}
\caption{The selected examples of sentiment control between our method and vanilla prompt tuning. We highlight the words relevant to the training corpus domain in red color. Vanilla prompt-tuning trends to overly imitate the feature of the training corpus (movie reviews) beyond sentiment attribute. However, our method could overcome this drawback well in most cases.}
\label{vanilla_example}
\end{table*}

\begin{table*}[h]

\renewcommand\arraystretch{1.25} 

\resizebox{\linewidth}{!}{


\begin{tabular}{ll}
\hline

\multicolumn{2}{l}{\textbf{Prompt:} He was accused of two offences, or doctrinal \textbf{(Positive Steering)}}                                                                                                                                        \\ \hline
PPLM                                & \begin{tabular}[c]{@{}l@{}}He was accused of two offences, or doctrinal crimes or revolt by doing:… \\ slapping the breasts of a virgin,  and she conceived by\end{tabular}                             \\
DART                                & \begin{tabular}[c]{@{}l@{}}He was accused of two offences, or doctrinal offences within religious circles,"\\ she added. "It's hard being here today on Easter this year because\end{tabular}                                \\
CTRL                                & \begin{tabular}[c]{@{}l@{}}He was accused of two offences, or doctrinal and spiritual heresies, for which\\  he was condemned by a church council in Rome. These charges are not mentioned\end{tabular} \\
GEDI                                & \begin{tabular}[c]{@{}l@{}}He was accused of two offences, or doctrinal  offenses, in regard to his \\ ordination to the ministry during the War of Independence of 1814\end{tabular}                   \\
DEXPERT                             & \begin{tabular}[c]{@{}l@{}}He was accused of two offences, or doctrinal breaches, relating to a prayer \\ campaign organised by Christians in northern Russia in 2008."He\end{tabular}                  \\
Vanilla Prompt                      & \begin{tabular}[c]{@{}l@{}}He was accused of two offences, or doctrinal sins, but the most disturbing \\ one was the one he committed against his own mother. The result is the\end{tabular}            \\
DisCup (Ours)                       & \begin{tabular}[c]{@{}l@{}}He was accused of two offences, or doctrinal offences; but the judge found \\ he had good reason to believe that he would not do this."\end{tabular}                         \\ \hline
\multicolumn{2}{l}{\textbf{Prompt:} UCSF is the nation's leading \textbf{(Negative Steering )}}                                                                                                                                                     \\ \hline

PPLM                                & \begin{tabular}[c]{@{}l@{}}UCSF is the nation's leading integrative university, offering countless students a world-class \\ educational experience in all subjects, from pre\end{tabular}              \\
DART                                & \begin{tabular}[c]{@{}l@{}}UCSF is the nation's leading health care research, teaching and patient\\  care organization. In fiscal year 2015, it directly funded and assisted\end{tabular}              \\
CTRL                                & \begin{tabular}[c]{@{}l@{}}UCSF is the nation's leading provider of online student assistance and \\ services. For further information, contact us.http://m.theage.com.au/sport/motor@@\end{tabular}    \\
GEDI                                & \begin{tabular}[c]{@{}l@{}}UCSF is the nation's leading  supporter of hospitals.  I'm not trying \\ to be disingenuous, but after last night's hurtful\end{tabular}                                     \\
DEXPERT                             & \begin{tabular}[c]{@{}l@{}}UCSF is the nation's leading provider of undergraduate and graduate \\ degrees in college-level psychology.  A subsidiary, University of California–San\end{tabular}         \\
Vanilla Prompt                      & \begin{tabular}[c]{@{}l@{}}UCSF is the nation's leading student newspaper, and its staff is a motley\\  crew of left-leaning intellectuals and liberal activists. But\end{tabular}                      \\
DisCup (Ours)                                & \begin{tabular}[c]{@{}l@{}}UCSF is the nation's leading provider in healthcare, but only 3.3\% of students \\ receive any healthcare assistance  in the first year or less\end{tabular}                 \\ \hline
\end{tabular}

}

\caption{The generation examples of sentiment control over some CTG approaches. We chose two challenging prompts with opposite polarity compared to the target sentiment. }
\label{sentiment_example}
\end{table*}

\begin{table*}[]
\resizebox{\linewidth}{!}{
\begin{tabular}{ll}
\hline
\multicolumn{2}{l}{\textbf{Prompt:} The prisoners are the bad guys, and}                                                                                                                                \\ \hline
PPLM           & \begin{tabular}[c]{@{}l@{}}The prisoners are the bad guys, and herson is Peter, the evil one. And to soothe his pain,\\  he is going to\end{tabular}                          \\
DART           & \begin{tabular}[c]{@{}l@{}}The prisoners are the bad guys, and we all have to work together to destroy the bad guys\\  and prevent them from further disasters,”\end{tabular} \\
GEDI           & \begin{tabular}[c]{@{}l@{}}The prisoners are the bad guys, and I was into the struggle for their freedom. \\ At school in Thailand's Narathiwat prison,\end{tabular}          \\
DEXPERT        & \begin{tabular}[c]{@{}l@{}}The prisoners are the bad guys, and  by setting their cages up so close to the street,\\  they were the least of their problems. But\end{tabular}  \\
Vanilla prompt & \begin{tabular}[c]{@{}l@{}}The prisoners are the bad guys, and the good guys are in jail. The bad guys will always \\  get a free pass. The prisoners\end{tabular}             \\
DisCup (Ours)   & \begin{tabular}[c]{@{}l@{}}The prisoners are the bad guys, and the prisoners can get away with it, but it's not the end. \\ We can get more justice if\end{tabular}           \\ \hline
\end{tabular}}
\caption{The generation examples of toxicity avoidance over different CTG methods.}
\label{detoxic_example}
\end{table*}

\label{sec:appendix}